\pdfoutput=1

\documentclass[letterpaper, 10 pt, conference]{ieeeconf}  

\usepackage{hyperref}
\usepackage{url}
\usepackage{graphicx}
\usepackage[utf8]{inputenc}
\usepackage{amsmath,amsfonts}
\usepackage{amssymb}
\usepackage{subfigure}
\usepackage{multirow}
\usepackage{multicol}
\usepackage{tikz}
\usepackage{xcolor,soul,color}
\makeatletter
\renewcommand*\env@matrix[1][*\c@MaxMatrixCols c]{%
  \hskip -\arraycolsep
  \let\@ifnextchar\new@ifnextchar
  \array{#1}}

\IEEEoverridecommandlockouts                              

\newcommand\copyrighttext{%
  \footnotesize This work has been submitted to the IEEE for possible publication. Copyright may be transferred without notice, after which this version may no longer be accessible.}
  
\newcommand\copyrightnotice{%
\begin{tikzpicture}[remember picture,overlay]
\node[anchor=south,yshift=10pt] at (current page.south) {\fbox{\parbox{\dimexpr\textwidth-\fboxsep-\fboxrule\relax}{\copyrighttext}}};
\end{tikzpicture}%
}

\title{\LARGE \bf
Relative ultra-wideband based localization of multi-robot systems with kinematic extended Kalman filter
}

\author{Salma Ichekhlef, Étienne Villemure, Shokoufeh Naderi, François Ferland and Maude Blondin
\thanks{All authors are with the Department of Electrical Engineering and Computer Engineering of the Université de Sherbrooke (2500 boulevard de l'Université, Sherbrooke, Québec (Canada), J1K-2R1). F. Ferland and M. Blondin are IEEE members. Contacts for the authors:
\{Etienne.Villemure, Salma.Ichekhlef, Shokoufeh.Naderi,  Francois.Ferland, and Maude.Blondin2\}@usherbrooke.ca.} 
}

\begin{document}

\maketitle
\thispagestyle{empty}
\pagestyle{empty}

\begin{abstract}

Localization plays a critical role in the field of distributed swarm robotics. Previous work has highlighted the potential of relative localization for position tracking in multi-robot systems. Ultra-wideband (UWB) technology provides a good estimation of the relative position between robots but suffers from some limitations. This paper proposes improving the relative localization functionality developed in our previous work, which is based on UWB technology. Our new approach merges UWB telemetry and kinematic model into an extended Kalman filter to properly track the relative position of robots. We performed a simulation and validated the improvements in relative distance and angle accuracy for the proposed approach. An additional analysis was conducted to observe the increase in performance when the robots share their control inputs.

\end{abstract}

\copyrightnotice

\section{INTRODUCTION}

Swarm robotics orchestrated with swarm intelligence requires high coordination, cooperation, and collaboration of multi-robot systems to perform specific tasks \cite{c1}. In this perspective, robot localization represents a significant problem to be solved, requiring first a powerful self-estimation of the location of each robot and second, a knowledge of the relative position of other robots~\cite{c2}.

The concept of localization encompasses two main aspects~\cite{c3}: absolute and relative localization. Absolute localization describes the position of robots in a global coordinate frame; in other words, it allows positioning based on a fixed point. Relative localization, on the other hand, is concerned with the localization of the relative configurations of mobile robots in relation to other robots or landmarks and does not have a fixed reference. The Global Positioning System (GPS) represents a typical solution that can perform both classes of localization (absolute and relative) \cite{c4}. However, this solution has several limitations, and it only works in environments with GPS reception, such as indoor and submarine deployments \cite{c5}.

Indeed, for multi-robot systems performing swarm robotics tasks, relative localization plays a crucial role in tracking a robot's self, and other robots' positions \cite{c3}. This localization concept is a critical operation that must be well mastered in complex environments featuring inaccurate and contaminated data. Therefore, it is necessary to find an error-tolerant solution to improve the accuracy, precision, and robustness of relative localization.

Relative localization has recently gained more attention in high-speed, low-power wireless communications due to the fine temporal resolution of ultra-wideband (UWB) radio technology, and it can provide good accuracy in relative location estimation, and tracking \cite{c6}. Moreover, UWB technology is based on operator-free sending and receiving of radio pulses using extremely-precise timing.


There are several techniques for measuring telemetry UWB, providing distances or relative angles between robots:

\begin{itemize}
\item Time of Arrival (ToA);
\item Time Difference of Arrival (TDoA);
\item Time of Flight (ToF);
\item Angle of Arrival or Phase Difference of Arrival (PDoA);
\item Received Signal Strength Indication (RSSI).
\end{itemize}

A review of these techniques shows they suffer from problems such as reflections, multipath, and non-line-of-sight measurements (NLOS) \cite{c7}. For this reason, combining the UWB technique with a state estimator is essential to achieve better performance.

In addition, an often used algorithm for sensor estimation are based on the Kalman filter, which provides real-time updates of a linear system's position. A more optimal solution for nonlinear systems is the extended Kalman filter (EKF)~\cite{c2, c6}.

In this paper, we propose improving the relative localization accuracy of multi-robot systems. To do so, we first measure the relative distance and orientation angle between mobile robots measured from the two UWB telemetry measurement techniques of ToF and PDoA, respectively. Then, we merge the different information obtained from these two techniques using EKF. The main contributions of this paper are :
\begin{enumerate}
    \item The EKF uses the kinematic model of robots as opposed to most of the other works that use the dynamic model. As a result, it eliminates the need to provide the robots' mass and moment of inertia to the estimator.
    \item An analysis of the performance degradation of the proposed EKF when it receives information from other robots at a lower frequency than the estimator's frequency. This analyses the sensitivity of the system to communication degradation. 
\end{enumerate}

The remainder of this paper is organized as follows: 
\hyperref[sec2]{Sect. \ref{sec2}} presents the related work. \hyperref[sec3]{Sect. \ref{sec3}} focuses on the relative position estimation approach, beginning with the problem statement and the description of the system model. Then, we detail the UWB-based relative localization method and the kinematic model used in the EKF. The simulation results to validate the proposed approach are conducted in \hyperref[sec4]{Sect. \ref{sec4}}. Finally, the conclusion and future work are given in \hyperref[sec5]{Sect. \ref{sec5}}.

\section{RELATED WORK} \label{sec2}

Multi-robot localization is often based on maps and/or landmarks to generate a global coordinate system \cite{c8}.
%
Kurazume et al. \cite{c9} proposed a "cooperative positioning with multiple robots" method that divides robots into two groups, one of which remains stationary to serve as a temporary landmark for the other group and vice versa until reaching the target. However, this method does not allow all robots to move simultaneously.
Without preventing the simultaneous movement of robots, Howard et al. \cite{c10} developed an approach that allows each robot to estimate the relative position of all nearby robots and broadcast these positions to all other robots. Robots generate an egocentric estimate of other robots' positions using Bayesian formalism and a particle filter. However, using a particle filter generally requires more computational power compared to the Kalman filter.

The use of the Kalman filter appears in \cite{c2, c11}, which allows merging the position measurements of continuously moving robots with relative localization information from other robots to locate them more accurately.
However, the first approach uses lidar and wheel odometry and the second approach uses wheel odometry only. Those are different position measurement technologies compared to UWB-based telemetry that provides good accuracy in relative position estimation and tracking \cite{c6}.

The use of UWB telemetry is  presented in \cite{c5, c12} in combination with a particle filter. An approach very similar to ours is developed in \cite{c13} with additional sensors like an altimeter and a camera. However, our method differs by removing those mentioned sensors and with a reduced number of UWB nodes, thus saving costs.


\section{RELATIVE LOCALIZATION} \label{sec3}
To formulate the relative localization problem, we use the SwarmUS platform \cite{c14} that suffers from inaccuracies issues that could be improved with the proposed approach.

\subsection{System description}
\begin{figure}[thpb]
    \centering
    \includegraphics[scale=0.7]{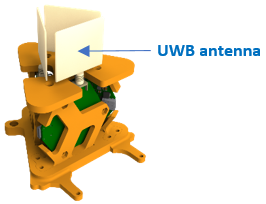}
    \caption{Three Beeboards assembly with antennas separated by 2.7 cm}
    \label{Fig1}
\end{figure}

The SwarmUS platform \cite{c14} is an embedded system that can be installed on robots to allow them to coordinate and communicate between themselves.
In addition to the main electronic board, called the Hiveboard, an assembly of three smaller boards (\hyperref[Fig1]{Fig. \ref{Fig1}}), called the Beeboards, can be added to a Hiveboard. Those Beeboards are UWB nodes that use the Decawave DW-1000 integrated circuit and have their own UWB antennas. 
They allow Hiveboards equipped with those assemblies to measure the relative distance and angle between each other.

The SwarmUS platform computes the relative distance using a two-way ranging (TWR) technique where three timestamped messages are exchanged between two robots. The distance is then computed from the time of flight measured from the timestamped messages.
The relative angle is computed using PDoA from the three different pairs of Beeboard that can be formed with three UWB nodes. Sequentially, one phase difference is measured, an angle of arrival per pair of antennas is then computed, and an estimated relative angle is produced for each pair before fusing them into a single angle measurement. 

\begin{figure}[bt]
    \centering
    \includegraphics[scale=0.65]{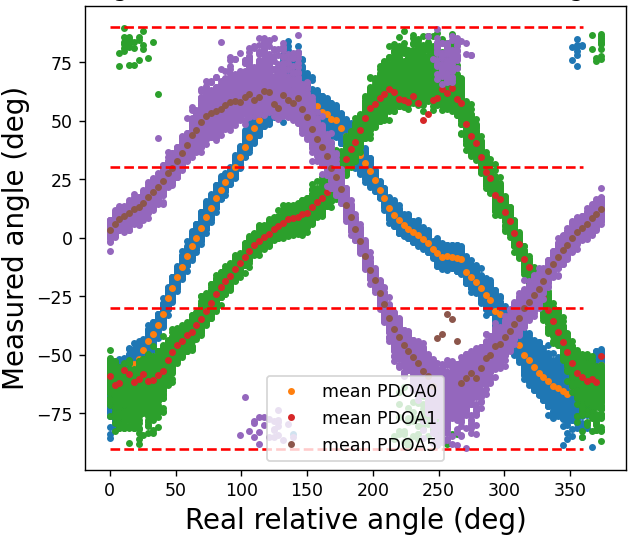}
    \caption{PDOA measured angles as the function of the real relative angle for the three antenna pairs}
    \label{FigPDOA}
\end{figure}

An example of the angles computed from each phase for each pair in the function of the real angle can be found at (\hyperref[FigPDOA]{Fig. \ref{FigPDOA}}). In this plot, 100 samples have been measured every 3.66$^{\circ}$ of the real angle from a Beeboard assembly on a rotating testbench while measuring a still Beeboard assembly two meters away without any obstructions. It can be observed that the dispersion of the samples around -90$^{\circ}$ and 90$^{\circ}$ are larger and some wrapping between those two extremities can occur, like shown around 125$^{\circ}$ and 260$^{\circ}$ in (\hyperref[FigPDOA]{Fig. \ref{FigPDOA}}). Those dispersions and wrapping create ambiguities in those areas and are a significant cause of inaccuracies in the system. It emphasizes furthermore the need for an additional modality to estimate the real angle.


\begin{figure}[b]
    \centering
    \includegraphics[scale=0.7]{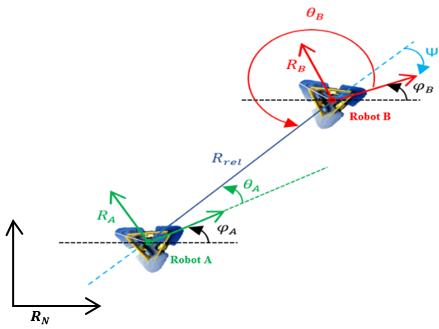}
    \caption{Presentation of the two-robot system}
    \label{Fig2}
\end{figure}

Our proposed approach relies on the same TWR and PDoA techniques, but with an EKF as an additional state estimator.
EKF allows the fusion of the information provided by the three UWB nodes and the kinematics of the system in order to provide a relative position closer to reality.
We reduce the problem to a system with two moving robots (\hyperref[Fig2]{Fig. \ref{Fig2}}). For convenience, the estimator is running on robot A and is tracking robot B. Each robot is equipped with three UWB nodes to provide telemetry measurements from the other robot, an inertial measurement unit (IMU) to describe its angular velocity, and wheel encoders to measure its linear velocity.

As shown in \hyperref[Fig2]{Fig. \ref{Fig2}}, $R_{rel}$ is the relative distance between the two robots. $\theta_A$ and $\theta_B$ are the orientation angles of robot A with respect to robot B and robot B with respect to robot A, respectively. Thus, $\phi_A$ and $\phi_B$ represent the rotation angles of the reference frame $R_A$ connected to robot A and the reference frame $R_B$ connected to robot B with respect to the absolute reference frame $R_N$, respectively.

In this paper, we perform relative localization (estimation of $R_{rel}$ and $\theta_A$) in 2 dimensions (2D) and assume that the three Beeboards antennas are placed in the center of each robot and that the angular and linear velocities are perfect and synchronized.

\subsection{Problem formulation}
To improve the accuracy of the relative localization, we introduce an EKF that fuses the telemetry measurements received from the three UWB nodes by integrating the kinematic described by the IMU and the robot's odometry.

A kinematic model is used instead of a dynamic one to remove the need to provide the robots' mass and moment of inertia to the estimator's model.
Using a kinematic model simplifies the configuration of the system on physically-different robots.

For each robot $i \in \{A, B\}$, we express its motion by the following equation:
\begin{equation} \label{eq1}
    \begin{bmatrix}
        \dot x_{i,k} \\
        \dot y_{i,k} \\
        \dot \phi_{i,k}
    \end{bmatrix}
    =
    \begin{bmatrix}
        \cos{\phi_{i,k}} & 0 \\
        \sin{\phi_{i,k}} & 0 \\
        0 & 1
    \end{bmatrix}
    \times
    \begin{bmatrix}
        v_{i,k} \\
        \dot{\phi}_{i,k}
    \end{bmatrix},
\end{equation}
where $[x_{i,k}\; y_{i,k}\; \phi_{i,k}]$ represents the absolute position of robot $i$ at time $k$, and $v_{i,k}$ represents the linear velocity of robot $i$ at time $k$.

To find the equations of $R_{rel}$ and $\theta_A$, we express the coordinates $[x_{B,k}^A \; y_{B,k}^A]$ of robot B in the $R_A$ frame as follows:
\begin{equation} \label{eq2}
    \begin{bmatrix}
        x_{B,k}^A \\
        y_{B,k}^A \\
        1
    \end{bmatrix}
    = T_{N,k}^A \times
    \begin{bmatrix}
        x_{B,k}^N \\
        y_{B,k}^N \\
        1
    \end{bmatrix},
\end{equation}

\begin{equation} \label{eq3}
    T_{N}^A =
    \begin{bmatrix}
        \cos{\phi_A} & \sin{\phi_A} & -x_A\cos{\phi_A}-y_A\sin{\phi_A} \\
        -\sin{\phi_A} & \cos{\phi_A} &  x_A\sin{\phi_A}-y_A\cos{\phi_A} \\
        0 & 0 & 1
    \end{bmatrix},
\end{equation}
where $T_{N}^A$ represents the transformation matrix from the $R_N$ frame to the $R_A$ frame.

From Fig. 2, \eqref{eq1}, and \eqref{eq2}, the relative position is estimated at each time $k$ as follows:
\begin{equation} \label{eq4}
    R_{rel,k} = \sqrt{( x_{A,k}^N-x_{B,k}^N )^2 + ( y_{A,k}^N-y_{B,k}^N )^2}\, ,
\end{equation}
\begin{equation} \label{eq5}
    \theta_{A,k} = \arctan2 \, (y_{B,k}^A, y_{B,k}^A)
\end{equation}

The kinematic model developed from equations \eqref{eq1} to \eqref{eq5} will be used to derive the estimator's equations but will also serve as the ground truth during the validation process to compare the different algorithms.

Since states of another robot relative to robot A appear in equations \eqref{eq4} and \eqref{eq5}, those states must be shared with the robot running the estimator.
They can be communicated by radio like the one used on the SwarmUS platform.

However, an increasing number of robots would lead to a bigger portion of the bandwidth being allocated to localization estimation.
This emphasizes the observation in our previous work~\cite{c14} where increasing the number of robots reduces the available bandwidth of each robot for its coordination and information sharing.
Thus, the information communicated by robot B might be transmitted at a lower frequency if the number of robots increases, the bandwidth usage is high or the radio signals are degraded. 
From this perspective, our method will treat two cases:
\begin{itemize}
\item \textbf{Case 1: } the EKF in robot A has access to information from robot B using the communication channel of SwarmUS. Validation of the estimator with different update frequencies from robot B will be analyzed.
\item \textbf{Case 2: } only the information of the robot A running the estimator is known. This will showcase if the estimator could afford to discard the communication bandwidth altogether and obtain better performance than the original algorithm.
\end{itemize}

\subsection{Mathematical model of relative localization}

\begin{figure}[t]
    \centering
    \includegraphics[scale=1.2]{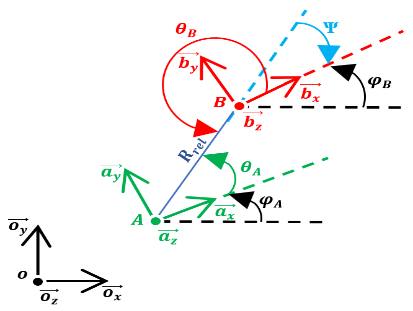}
    \caption{Illustration of the relative location device}
    \label{Fig3}
\end{figure}

We determine the expression for the time derivative of the relative position of robot A with respect to robot B ($\dot{R}_{rel}$ and $\dot{\theta}_A$). For simplicity, we model robot A and robot B by points A and B, respectively (\hyperref[Fig3]{Fig. \ref{Fig3}}).

We begin by expressing the time derivative in the $R_N$ frame of center $O$ and base $(\overrightarrow{o_x}, \overrightarrow{o_y}, \overrightarrow{o_z})$ of the vector $\overrightarrow{AB}$ by its coordinates in the $R_A$ frame of center $A$ and base $(\overrightarrow{a_x}, \overrightarrow{a_y}, \overrightarrow{a_z})$ (Fig. 3). According to Bour's formula, we obtain the following equation:
\begin{equation} \label{eq6}
    \frac{d\overrightarrow{AB}}{dt}_{R_N} = \frac{d\overrightarrow{AB}}{dt}_{R_A} + \overrightarrow{\Omega}_{R_A/R_N} \cap \overrightarrow{AB} \, ,
\end{equation}
with
\begin{equation} \label{eq7}
    \begin{cases}
        \overrightarrow{AB} = R_{rel} \cos{\theta_A} \overrightarrow{a_x} + R_{rel} \sin{\theta_A} \overrightarrow{a_y}\\
        \overrightarrow{\Omega}_{R_A/R_N} = \dot{\phi}_A \overrightarrow{a_z} \, ,
    \end{cases}
\end{equation}
where $\overrightarrow{\Omega}_{R_A/R_N}$ represents the angular velocity of the $R_A$ frame relative to the $R_N$ frame.

\begin{equation} \label{eq8}
    \frac{d\overrightarrow{AB}}{dt_{R_N}} = 
    \begin{matrix}[c|c]
        \overrightarrow{a_x} & \dot{R}_{rel} \cos{\theta_A} - R_{rel} \dot{\theta}_A \sin{\theta_A} - R_{rel} \dot{\phi}_A \sin{\theta_A}\\
        \overrightarrow{a_x} & \dot{R}_{rel} \sin{\theta_A} + R_{rel} \dot{\theta}_A \cos{\theta_A} + R_{rel} \dot{\phi}_A \cos{\theta_A}\\
        \overrightarrow{a_x} & 0 \\
    \end{matrix}
\end{equation}

We also have:
\begin{equation} \label{eq9}
    \frac{d\overrightarrow{AB}}{dt}_{R_N} = \frac{d\overrightarrow{OB}}{dt}_{R_A} - \frac{d\overrightarrow{OA}}{dt}_{R_A} \, .
\end{equation}

We express the two time derivatives $\frac{d\overrightarrow{OB}}{dt}_{R_A}$ and $\frac{d\overrightarrow{OA}}{dt}_{R_A}$ at the base $(\overrightarrow{a_x}, \overrightarrow{a_y}, \overrightarrow{a_z})$ by the following equations:
\begin{equation} \label{eq10}
    \begin{cases}
        \frac{d\overrightarrow{OA}}{dt}_{R_A} = R_N^A \times
        \begin{bmatrix}
            v_A \cos{\phi_A} \\
            v_A \sin{\phi_A} \\
            0
        \end{bmatrix}\\
        \frac{d\overrightarrow{OB}}{dt}_{R_A} = R_N^A \times
        \begin{bmatrix}
            v_A \cos{\phi_A} \\
            v_A \sin{\phi_A} \\
            0
        \end{bmatrix}\\
    \end{cases} \, ,
\end{equation}
\begin{equation} \label{eq11}
    R_N^A = 
    \begin{bmatrix}
        \cos{\phi_A} & -\sin{\phi_A} \\
        \sin{\phi_A} & \cos{\phi_A} \\
    \end{bmatrix} \, ,
\end{equation}
where $R_N^A$ represents the rotation matrix from frame $R_N$ to frame $R_A$. Thus, $v_A$ and $v_B$ are the 
linear velocities
of robot A and robot B, respectively.

From \eqref{eq9}, \eqref{eq10}, and \eqref{eq11}, we obtain a second expression of $\frac{d\overrightarrow{AB}}{dt}_{R_N}$:
\begin{equation} \label{eq12}
    \frac{d\overrightarrow{AB}}{dt}_{R_N} = 
    \begin{matrix}[c|c]
        \overrightarrow{a_x} & v_B \cos{\phi_B-\phi_A} - v_A\\
        \overrightarrow{a_x} & v_B \sin{\phi_B-\phi_A}\\
        \overrightarrow{a_x} & 0 \\
    \end{matrix}
    \, .
\end{equation}

Then, we equalize the \eqref{eq8} and \eqref{eq12}. After doing the calculation, we obtain \eqref{eq13}, which models the temporal variation of the relative position of robot A with respect to robot B:
\begin{equation} \label{eq13}
    \begin{cases}
        \dot{\theta}_A = -\dot{\phi}_A - \frac{v_B \sin{\theta_A + \phi_A - \phi_B} - v_A \sin{\theta_A}}{R_{rel}} \\
        \dot{R}_{rel} = v_B \cos{\theta_A + \phi_A - \phi_B} - v_A \cos{\theta_A} \\
    \end{cases} \, .
\end{equation}

From Fig. 3, we can see that:
\begin{equation} \label{eq14}
    \theta_B = \theta_A + \phi_A - \phi_B + \pi \, .
\end{equation}

From \eqref{eq13} and \eqref{eq14} we derive a simplified expression for the time variation of the relative position:
\begin{equation} \label{eq15}
    \begin{cases}
        \dot{\theta}_A = -\dot{\phi}_A + \frac{v_B \sin{\theta_B} + v_A \sin{\theta_A}}{R_{rel}} \\\dot{R}_{rel} = -(v_B \cos{\theta_B} + v_A \cos{\theta_A}) \\
    \end{cases} \, .
\end{equation}

\subsection{Description of the EKF formulation} \label{subsec.IIID}

\textbf{Formulation of EKF:} The state vector of our EKF describes the relative position of the two-robot system at each time $k$ and is denoted by $X_k$:
\begin{equation} \label{eq16}
    X_k = 
    {\begin{bmatrix}
        \theta_{A,k} & R_{rel,k}
    \end{bmatrix}}^T \, .
\end{equation}

At each time $k$, the three active Beeboard antennas of robot A allow us to calculate the ToF and the three PDoA~\cite{c15} with respect to robot B. This information will constitute the measured outputs $Y_k$ of our EKF, such that:
\begin{equation} \label{eq17}
    Y_k = 
    {\begin{bmatrix}
        \theta_{1_m,k} & \theta_{2_m,k} & \theta_{3_m,k} & R_{rel_m,k}
    \end{bmatrix}}^T \, ,
\end{equation}
where $R_{rel_m,k}$ and $\theta_{1_m,k} (i \in \{1,2,3\})$ represent the relative distance between the two robots and the PDoA of the antenna pair $i$ measured at time $k$, respectively.

In addition, the information obtained from the IMU sensors, the wheel's odometry, and the relative orientation angle estimated by the other robot $\theta_B$ are considered inputs to our EKF. We denote it by a control vector $U_k$.

The state transition and observation models ($F(X_k, U_k)$ and $H(X_k)$, respectively) of our system are nonlinear functions of the state:
\begin{equation} \label{eq18}
    X_k = F(X_{k-1}, U_k) + W_k \, ,
\end{equation}
\begin{equation} \label{eq19}
    Y_k = H(X_k) + V_k \, ,
\end{equation}
where $W_k$ and $V_k$ are the process and observation noises. We assume that these noises are Gaussian with zero means and covariances $Q$ and $R_k$, respectively.
Since each pair of antennas in the setup  has a dispersion that depends on $\theta_{A}$, the $R_k$ matrix will change at each step in function of the estimated $\theta_{A}$.
The variance changes based on a lookup table for each antenna pair containing standard deviations computed from experimental data as in \hyperref[FigPDOA]{Fig. \ref{FigPDOA}}.


\textbf{Prediction:} In this step, the EKF estimates the expected relative position $\hat{X}_{k|k-1}$ and error covariance $P_{k|k-1}$:
\begin{equation} \label{eq20}
    \hat{X}_{k|k-1} = F(\hat{X}_{k-1|k-1}, U_k) \, ,
\end{equation}
\begin{equation} \label{eq21}
    P_{k|k-1} = F_k P_{k-1|k-1} F_k^T + Q \, ,
\end{equation}
where $\hat{X}_{k|k-1}$ and $P_{k|k-1}$ are the updates of the estimation of the state and error covariance at time $k-1$, respectively. Thus, $F_k$ is the transition Jacobian matrix.

\textbf{Update:} At this step, the EKF updates the relative position estimate $\hat{X}_{k|k}$ and the error covariance $P_{k|k}$:
\begin{equation} \label{eq22}
    \hat{X}_{k|k} = \hat{X}_{k|k-1} + K_k(Y_k - H\hat{X}_{k|k-1}) \, ,
\end{equation}
\begin{equation} \label{eq23}
    P_{k|k} = (I - H_k K_k) P_{k|k-1} \, ,
\end{equation}
\begin{equation} \label{eq24}
    K_k = P_{k|k-1} H_k^T {(H_k P_{k|k-1} H_k^T + R_k)}^{-1} \, ,
\end{equation}
where $K_k$ is the Kalman gain vector and $H_k$ is the observation Jacobian matrix.

As mentioned in the problem statement section, our method will deal with two cases.

\textbf{Case 1:} In this case, the control vector contains the following entries:
\begin{equation} \label{eq25}
    U_k = [v_{A,k} \; \dot{\phi}_{A,k} \; v_{B,k} \; \theta_{B,k}]^T\, ,
\end{equation}
where $v_{A,k}$ and $v_{B,k}$ are the velocities of robot A and robot B at time $k$, respectively. $\dot{\phi}_{A,k}$ represents the rotational velocity of robot A at time $k$, and $\theta_{B,k}$ is the relative orientation angle estimated at time $k$ by robot B.

From \eqref{eq15}, \eqref{eq16}, \eqref{eq18} and \eqref{eq25}:
\begin{align} \label{eq26}
    &F(X_{k-1}, U_k) = X_{k-1} + \\ 
    &\Delta T
    \begin{bmatrix}
        -U_k[2] + \frac{U_k[1] \sin{X_{k-1}[1]} + U_k[3] \sin{U_k[4]}}{X_{k-1}[2]}\\
        -U_k[1] \cos{X_{k-1}[1]} - U_k[3] \cos{U_k[3]}\\
    \end{bmatrix} \, , \notag
\end{align}
and
\begin{align} \label{eq27}
    &F_k = I_2 + \\ 
    &\Delta T
    \begin{bmatrix}
        \frac{U_k[1]}{X_{k-1}[2]} \cos{X_{k-1}[1]} & -\frac{U_k[1] \sin{X_{k-1}[1]} + U_k[3] \sin{U_k[4]}}{(X_{k-1}[2])^2}\\
        U_k[1] \sin{X_{k-1}[1]} & 0\\
    \end{bmatrix} \, , \notag
\end{align}
where $\Delta T$ represents the duration between two successive states.

From \eqref{eq16}, \eqref{eq17}, and \eqref{eq19}, we have:
\begin{equation} \label{eq28}
    H(X_k) = 
    \begin{bmatrix}
        A_{1,k}X_k[1]+B_{1,k}\\
        A_{2,k}X_k[1]+B_{2,k}\\
        A_{3,k}X_k[1]+B_{3,k}\\
        X_k[2]
    \end{bmatrix} \, ,
\end{equation}
where $A_{i,k}$ and $B_{i,k} \, (i \in \{1,2,3\})$ represent the coefficients describing the calibration \cite{c15} curves between the actual rotation angle and PDoA of each antenna pair $i$.

We also have:
\begin{equation} \label{eq29}
    H(k) = 
    \begin{bmatrix}
        A_{1,k} & 0\\
        A_{2,k} & 0\\
        A_{3,k} & 0\\
        0 & 1
    \end{bmatrix} \, ,
\end{equation}

\eqref{eq28} and \eqref{eq29} are valid for both cases.

\textbf{Case 2:} In this case, the control vector contains the following entries:
\begin{equation} \label{eq30}
    U_k = [v_{A,k} \; \dot{\phi}_{A,k}]^T\, ,
\end{equation}

From \eqref{eq15}, \eqref{eq16}, \eqref{eq18} and \eqref{eq30}, we have:
\begin{equation} \label{eq31}
    F(X_{k-1}, U_k) = X_{k-1} + \Delta T
    \begin{bmatrix}
        -U_k[2] + \frac{U_k[1] \sin{X_{k-1}[1]}}{X_{k-1}[2]}\\
        -U_k[1] \cos{X_{k-1}[1]}\\
    \end{bmatrix} \, ,
\end{equation}
and
\begin{equation} \label{eq32}
    F_k = I_2 + \Delta T
    \begin{bmatrix}
        \frac{U_k[1]}{X_{k-1}[2]} \cos{X_{k-1}[1]} & -\frac{U_k[1] \sin{X_{k-1}[1]} }{(X_{k-1}[2])^2}\\
        U_k[1] \sin{X_{k-1}[1]} & 0\\
    \end{bmatrix} \, .
\end{equation}
 For both cases, it must be noted that, since $\theta_A$ is a bearing angle, it wraps between 0$^{\circ}$ and 360$^{\circ}$.
Thus, the output of the prediction steps for the angle is reduced by modulo 360 to properly take into account the wrap.

\section{SIMULATION AND RESULTS} \label{sec4}
To validate the proposed approach, a Simulink model was created to run the original system in parallel with the EKF model presented. They all estimate the relative distance and angle of a moving robot B from a moving robot A. In this section, we first describe the simulator, the performed tests, and their results.

\subsection{Description of the simulator}
The Simulink models simulate the kinematics of the robots, the noisy measurements of the UWB nodes, the process of the SwarmUS system to generate angles from phases, and all the compared estimators. An overview of this simulator architecture is given in \hyperref[Fig34]{Fig. \ref{Fig34}}.
\subsubsection{Robot physics systems}
The blue blocks in \hyperref[Fig34]{Fig. \ref{Fig34}} simulate the physics of the robot based on the integration of \eqref{eq1} and compute the real relative distance and angle between the two robots with \eqref{eq4} and \eqref{eq5}. The inputs of those blocks are predefined time series vectors of $U_k$. Those inputs and the initial positions of the robots change the simulated trajectories of the robot. They are defined in a Matlab script based on scenarios that need to be tested. 

\subsubsection{Measurement simulation of the UWB nodes} \label{measurement_sim_UWB}
The yellow blocks in \hyperref[Fig34]{Fig. \ref{Fig34}} simulates how the angles of arrival are perceived by the UWB nodes. The main goal of this segment is to add realistic noise to the measurements.
For the distance, we inject a noise with a standard deviation of 3.43~cm, which corresponds to our previous observations~\cite{c14}.

For the relative angle, we simulate what each UWB pair will measure as their angle of arrival by generating a random value from the normal distribution that characterized each pair at that real relative angle. Those distributions are stored in a lookup table for each pair that contains non-zero mean distributions from 0$^{\circ}$ to 360$^{\circ}$. Those distributions were computed from a dataset similar to the one shown in \hyperref[FigPDOA]{Fig. \ref{FigPDOA}} with 100 samples every 3.66$^{\circ}$ with a Beeboard assembly on a rotating test bench.
The normality of the distribution was validated with a Shapiro-Wilk normality test~\cite{c15}.

\subsubsection{Emulation of the SwarmUS localization}
\begin{figure*}[htb]
    \centering
    \includegraphics[scale=0.6]{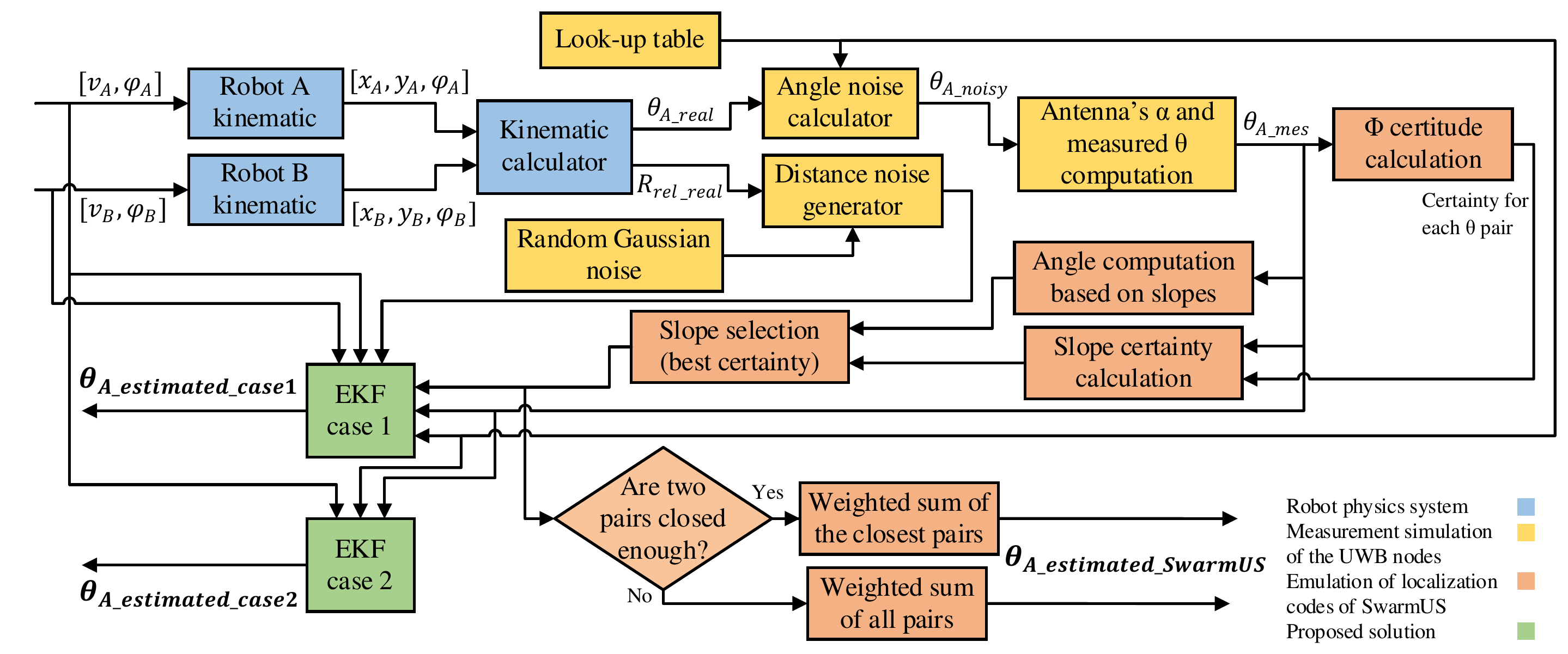}
    \caption{Block diagram of the simulation}
    \label{Fig34}
\end{figure*}

The orange blocks in \hyperref[Fig34]{Fig. \ref{Fig34}} emulates the embedded code that runs inside the SwarmUS hardware. This process contains the original algorithm that we use for comparison to the EKFs. A detailed description can be found in our documentation~\cite{c15}.

The algorithm goes through four main steps after receiving the angle of arrivals.
First, for each pair a certainty value is computed for each measured PDoA and we determine six possible relative angles generated from the two calibration slopes linked to each pair. The second step is to use the values of the three measured PDoAs with their certainty to calculate the certainty of each calibration slope (rising and failling) for each antenna pair. In the third step, based on the different calculated certainties of each slope, we reduce the six possible relative angles to three relative angles by selecting angles related to the slopes with the highest certainty.  The last step consists of combining the two closest angles together if they are close enough or otherwise the three estimated angles into one final angle. After estimating the relative angle, we use an exponential moving average filter to eliminate the less significant fluctuations.
This last angle is the output of the original algorithm.

\subsubsection{EKFs (proposed solutions)}
All the EKFs' formulations presented in section \ref{subsec.IIID} are integrated into two Simulink's EKF blocs shown in green in \hyperref[Fig34]{Fig. \ref{Fig34}}.
Since Case 1's EKF has a complete model of the system, it can trust more its model compared to the incomplete model of Case 2's EKF. Thus, their Q matrix linked to their process noise are two diagonal matrices with $10^{-6}$ and $10^{-3}$ terms, respectively. 

Since the dispersion of the angles of arrival for each pair is known at a specified angle using the lookup table ,
the R matrices of the EKF are adapted at each time iteration with an estimated variance for all its measurements. The estimated  $\theta_{A}$ is used to consult the lookup table.

As shown in equation \eqref{eq28}, the measurement process of both EKFs needs the coefficients of the selected slopes to compute their outputs. Therefore, the selected slopes of the SwarmUS code are fed into the estimators.

Since we are interested in the steady state operation of the EKFs and not in the convergence phase of the filter, we set the initial states of the EKFs to their real states.

\subsubsection{Timing and signal rate}
The simulation lasts 20~s with a fixed time step of 1~ms which is approximately the rate at which SwarmUS can process a UWB measurement.
However, the EKFs are executing at a rate of 28 Hz which is the maximum rate at which two robots in turn localize each other using SwarmUS~\cite{c14}. All the incoming inputs are then fed at this same rate since the robots could also provide speed measurements at this frequency. Zero-hold blocks are used to simulate this behavior in Matlab.

Since the rate of the incoming information from the robot in $U_K$ might change for Case 1, a down-sampled zero-hold block is used to simulate a lower data rate.

\subsection{Performed tests}
The original solution and the proposed estimator were tested in different motion scenarios to cover a broad range of values for the $\theta_{A}$ and $R_{rel}$. For this purpose, 14 different motion scenarios combining different linear velocities, angular velocities, acceleration, and initial positions have been implemented creating motion from robots standing still to both spiraling around each other. 
For each scenario, 10 simulations were run varying a parameter of its underlying motion.
To analyze the influence of the varying data rate from robot B for Case 1, all those 140 simulations were run at different incoming frequencies of $v_{B}$ and $\theta_{B}$. We set this data rate to be the 28~Hz frequency used by the EKF divided by an integer going from 1 to 10. Thus the performance of the EKF will be analyzed when information from the robot comes at the full data rate of 28 Hz down to 2.8~Hz.

The root-mean-square error (RMSE) is used as a performance metric for all the estimated states against their ground truth value for the relative distance and angle.

\begin{table}[hb!]
	\caption{Median of the RMSE of the 140 simulations' results}
	\begin{center}
		\begin{tabular}{cc p{1.2cm}p{1.2cm}c}
            \hline 
			\rule{0pt}{4mm}%
			Estimation source & Original & Case 1 (fb = 28Hz) & Case 1 (fb = 2.8 Hz) & Case 2  \\[1mm] \hline
            \rule{0pt}{4mm}%
			Distance (m)         & 0.0324   & 0.0207             & 0.0399               & 0.0265  \\
            Angle (deg)           & 16.0422  & 1.8267             & 4.3997               & 15.8068 %
            \\[1mm]  \hline
		\end{tabular}
	\end{center}
	\label{table1}
\end{table}

\begin{table*}[htb!]
	\caption{$p-$values of Wilcoxon ranked test between the original solution and the proposed approach}
	\begin{center}
		\begin{tabular}{cccccccccccc}
			\hline \\
			\multicolumn{2}{c}{Data rate B (Hz)} & 28 & 14 & 9.33 & 7 & 5.6 & 4.66 & 4 & 3.5 & 3.11 & 2.8 %
			\\[1mm]  \hline 
			\rule{0pt}{4mm}%
			\multirow{2}{*}{Distance} & Case 1 & 9.09e-7 & 0.001 & 0.009 & 0.028 & 0.131 & 0.219 & 0.374 & 0.410 & 0.556 & 0.807 \\
             & Case 2 & 8.32e-6 & 8.32e-6 & 8.32e-6 & 8.32e-6 & 8.32e-6 & 8.32e-6 & 8.32e-6 & 8.32e-6 & 8.32e-6 & 8.32e-6%
			\\[1mm] \hline
            \rule{0pt}{4mm}%
			\multirow{2}{*}{Angle} & Case 1 & 1.28e-20 & 6.51e-17 & 5.62e-16 & 6.09e-15 & 9.61e-14 & 1.32e-12 & 9.77e-12 & 2.46e-11 & 2.83e-11 & 2.62e-10 \\
             & Case 2 & 0.043 & 0.043 & 0.043 & 0.043 & 0.043 & 0.043 & 0.043 & 0.043 & 0.043 & 0.043 %
             \\[1mm] \hline
		\end{tabular}
	\end{center}
	\label{table2}
\end{table*}

\subsection{Simulation results}
\begin{figure}[tb]
    \centering
    \includegraphics[scale=0.6]{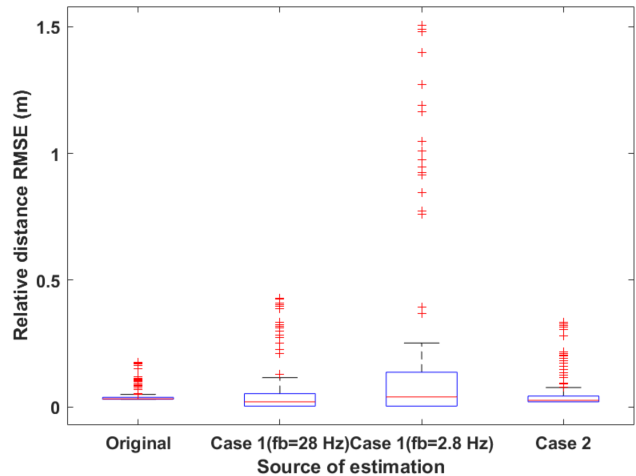}
    \caption{Boxplots of the RMSE of the distance of the 140 simulations}
    \label{FigRMSEdistance}
\end{figure}

\begin{figure}[tb]
    \centering
    \includegraphics[scale=0.6]{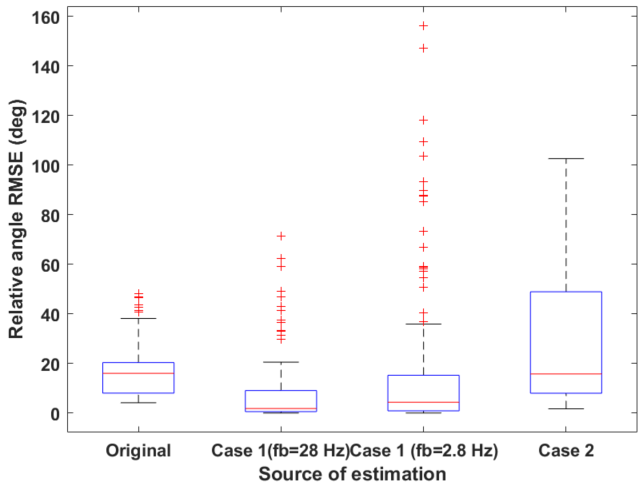}
    \caption{Boxplots of the RMSE of the angle of the 140 simulations}
    \label{FigRMSEangle}
\end{figure}

Boxplots of the RMSE of the distance and the angle for the original system, the EKF of Case 1 with a data rate of the robot at 28 Hz and 2.8 Hz, and the EKF of Case 2 are found at \hyperref[FigRMSEdistance]{Fig. \ref{FigRMSEdistance}} and \hyperref[FigRMSEangle]{Fig. \ref{FigRMSEangle}}, respectively. For clarity, the median of those boxplots is presented in Table~\ref{table1}. 

A Wilcoxon signed-rank test was performed to verify if the proposed approach differs significantly in performance with the orignal system.
The null hypothesis is that the median of both distributions are similar. The p-values of those tests performed between the original system and the EKFs for all robot B's data rates can be found  in Table \ref{table2}. We use an $\alpha=0.05$ to reject the null hypothesis.

A few observations can be made from \hyperref[FigRMSEdistance]{Fig. \ref{FigRMSEdistance}}, \hyperref[FigRMSEangle]{Fig. \ref{FigRMSEangle}}, Table \ref{table1} and  Table \ref{table1}. First, for all data rates of robot B, the statistical tests for both systems related to the angle reject the null hypothesis. Also, the medians of the EKF from Case 1 in the best and worst data rate situation and the EKF from Case 2 are lower than the median of the original system. Thus, the proposed  approach improves the performance of $\theta_A$ even when the data rate of information from robot B is degraded. Also, since both quantiles of the EKF with the complete model are almost both below the first quantile of the original system, results suggests that this improvement is very significant. Furthermore, the median is lowered to 1.8267$^\circ$ compared to the original 16.042$^\circ$. 

On the other hand, the statistical test shows that Case 1's EKF gives statistically different results until robot B's data rate is 7~Hz for the $R_{rel}$. Below that frequency, the median of the distance RMSE from this estimator could not be distinguished from the original system. So, with a sufficient data rate, the median of the distance could be improved, but the proposed approach seems to give worse result with a larger dispersion at lower data rates.
However, Case 2's EKF rejects the null hypothesis of the statistical test and has a lower median, thus showing an improvement in the distance measurements. Since this estimator has a Q matrix with higher values, it was expected that it will have results closer to the original system than Case 1 could.

However, even if we conclude that the estimators give improved estimations of the states most of the time, it also gives worst extreme values as it can be seen in \hyperref[FigRMSEdistance]{Fig. \ref{FigRMSEdistance}}, \hyperref[FigRMSEangle]{Fig. \ref{FigRMSEangle}}. These outliers appear mainly when the robots arrive at short relative distances, as was noticed at less than 30 cm. Since Case 1's EKF is affected the most, it suggests that the model is more sensitive in those scenarios and robustness to this phenomenon should be investigated in future work.

As final observations, it is important to notice that increasing the data rate of robot B's information improves the performance of the EKF by decreasing the median of the error data, narrowing its disperions and producing lesser outliers compared to not using any information from other robots. It also indicate that the use of a zero-hold mechanism to manage the low data rate problem was robust enough for the studied case.
However, other options such as adjusting the Q matrices in real time could lessen the disturbance of lower expected data rate.


\section{CONCLUSION AND FUTURE WORK} \label{sec5}
In this paper, we presented an UWB-based relative localization approach with improved distance and angle accuracy.
First, an estimation of the relative position between mobile robots is performed based on UWB telemetry measurement techniques.
Then, an EKF state estimator is used to correct these UWB telemetry measurements based on a kinematic model by using IMU and wheel odometry sensors.
Two cases are proposed based on data sharing availability between robots.
To validate the performance of our method, we performed a simulation that produced promising results in reducing the median error of the relative position estimate. Moreover, significatively better results were measured when enabling and increasing the data rate of information between the robots. Future work will be put towards increasing the robustness of the solution in specific motion scenarios and degraded communication channels, and validating the approach on a real physical system.

\addtolength{\textheight}{-11.2cm}   







\end{document}